\documentclass[a4paper,fleqn]{cas-dc}

\usepackage[numbers]{natbib}
\usepackage{amsmath,amsfonts}
\usepackage{algorithmic}
\usepackage{algorithm}
\renewcommand{\algorithmiccomment}[1]{\bgroup\hfill//~#1\egroup}
\usepackage{array}
\usepackage{textcomp}
\usepackage{stfloats}
\usepackage{verbatim}
\usepackage{graphicx}
\usepackage{amssymb}
\usepackage{booktabs}
\usepackage{multirow}
\usepackage[normalem]{ulem}
\useunder{\uline}{\ul}{}
\usepackage{commath}
\usepackage{adjustbox}
\usepackage{bookmark}
\usepackage{subcaption}

\def\methodUDSX{UDSX}
\def\methodUDSXFull{Unified Deep Semantic Expansion}
\def\methodDEX{DEX}

\def\DSDFull{Data Semantic Decoupling}
\def\DSDAbrv{DSD}
\def\PSTEFull{Progressive Spatio-Temporal Expansion}
\def\PSTEAbrv{PSTE}
\def\PTEFull{Progressive Temporal Expansion}
\def\PTEAbrv{PTE}
\def\ABSFull{Activation-Based Stratification}
\def\ABSAbrv{ABS}
\def\CSRFull{Contrastive-Stream Reunification}
\def\CSRAbrv{CSR}
\def\CSPFull{Contrastive-Stream Pairwise Loss}
\def\CSPAbrv{CSP}
\def\CSCFull{Contrastive-Stream Center Loss}
\def\CSCAbrv{CSC}
\def\CSTFull{Contrastive-Stream Triplet Loss}
\def\CSTAbrv{CST}
\def\Lagr{\mathcal{L}}

\def\tsc#1{\csdef{#1}{\textsc{\lowercase{#1}}\xspace}}
\tsc{WGM}
\tsc{QE}
\tsc{EP}
\tsc{PMS}
\tsc{BEC}
\tsc{DE}

\begin{document}
\begin{sloppypar}
\let\WriteBookmarks\relax
\def\floatpagepagefraction{1}
\def\textpagefraction{.001}

\shorttitle{A Unified Deep Semantic Expansion Framework for DG-ReID}

\shortauthors{Ang, Lin, Kot}

\title [mode = title]{A Unified Deep Semantic Expansion Framework for Domain-Generalized Person Re-identification}                      



%
\author{Eugene P.W. Ang}[type=editor,
                        auid=000,bioid=1,
                        orcid=0000-0001-7507-6569]
\ead{phuaywee001@e.ntu.edu.sg}
\credit{Conceptualization; Formal analysis; Investigation; Methodology; Software; Validation; Visualization; Roles/Writing - original draft; Writing - review \& editing}

\author{Shan Lin}[type=editor,
                        auid=001,bioid=2,
                        orcid=0000-0002-3254-1923]
\cormark[1]
\ead{shan.lin@ntu.edu.sg}
\credit{Methodology; Project administration; Resources; Supervision; Writing - review \& editing}

\author{Alex C. Kot}[type=editor,
                        auid=002,bioid=3,
                        orcid=0000-0001-6262-8125]
\ead{eackot@ntu.edu.sg}
\credit{Project administration; Resources; Supervision; Writing - review \& editing}






\affiliation{organization={Rapid-Rich Object Search (ROSE) Lab, Nanyang Technological University},
    addressline={School of EEE, 50 Nanyang Ave S2-B4b-13}, 
    postcode={639798}, 
    country={Singapore}}






\cortext[cor1]{Corresponding author}



\begin{abstract}
Supervised Person Re-identification (Person ReID) methods have achieved excellent performance when training and testing within one camera network. However, they usually suffer from considerable performance degradation when applied to different camera systems. In recent years, many Domain Adaptation Person ReID methods have been proposed, achieving impressive performance without requiring labeled data from the target domain. However, these approaches still need the unlabeled data of the target domain during the training process, making them impractical in many real-world scenarios. Our work focuses on the more practical Domain Generalized Person Re-identification (DG-ReID) problem. Given one or more source domains, it aims to learn a generalized model that can be applied to unseen target domains. One promising research direction in DG-ReID is the use of implicit deep semantic feature expansion, and our previous method, Domain Embedding Expansion (\methodDEX{}), is one such example that achieves powerful results in DG-ReID. However, in this work we show that \methodDEX{} and other similar implicit deep semantic feature expansion methods, due to limitations in their proposed loss function, fail to reach their full potential on large evaluation benchmarks as they have a tendency to saturate too early. Leveraging on this analysis, we propose \methodUDSXFull{}, our novel framework that unifies implicit and explicit semantic feature expansion techniques in a single framework to mitigate this early over-fitting and achieve a new state-of-the-art (SOTA) in all DG-ReID benchmarks. Further, we apply our method on more general image retrieval tasks, also surpassing the current SOTA in all of these benchmarks by wide margins.

\end{abstract}



\maketitle

\section{Introduction}
\label{sec:intro}
Person re-identification, also known as Person ReID, is the task of matching images or videos of the same person over a multi-camera surveillance system. Many fully supervised models already demonstrate impressive performance when trained and tested on the same dataset. However, most of these models often over-fit the training dataset (the source surveillance system) and usually suffer from considerable performance degradation when tested on unseen datasets (the target surveillance system). To address the large domain gap between source and target domains, many recent works utilize domain adaptation (DA) techniques to enhance the model's cross-domain capability. However, DA methods require target domain images/videos to adapt to the target system, which, in many real-world scenarios, is difficult to get. As a result, DA-ReID methods constrain the applicability of Person ReID and may delay system deployment. 

Our work in this paper follows the more practical Domain Generalization (DG) approach. Domain Generalized Person Re-identification, also known as DG-ReID, focuses on training a generalized model from multiple existing datasets without any prior knowledge of or access to the target domain. This is a more practical scenario for several reasons. Firstly, supervised or domain-adaptation methods require an extra round of retraining or adaptation when presented with a new domain. However, not all stakeholders are able to provide the resources for such retraining or adaptation as these methods take significant time and hardware to train~\cite{M3L, QAConv}. Furthermore, it may not be feasible to access data from the new domain because of privacy issues, necessitating that the method be unaware of the target domain~\cite{M3L}. Finally, for large scale commercialization it is ideal to offer solutions that can work out-of-the-box without extra fine-tuning or retraining, as not all stakeholders have the resources to do so~\cite{DIMN}.

Various techniques of DG-ReID such as meta-learning~\cite{M3L}, hyper-networks~\cite{DIMN} and memory banks~\cite{QAConv} have greatly advanced the state-of-the-art. Furthermore, a recent class of implicit semantic feature expansion methods has emerged~\cite{ISDA-TPAMI2020,ISDA-NeurIPS2019} that have shown promise in DG-ReID. Implicit semantic expansion methods emerge naturally as a resource-light alternative to explicit semantic expansion. Performing explicit semantic expansion on raw images and deep features~\cite{NIPS2017_RegularizingDeepNeuralNetworksbyNoise,he2019ParametricNoiseInjection,Lim2022NoisyFM} injects more semantic variety to a fixed dataset of images and can be viewed as a form of data augmentation. However, instead of explicitly manipulating the raw image or feature, implicit semantic methods encode this explicit semantic expansion into differentiable loss functions. As these loss functions are theoretical constructs, they are intractable to compute. Thus, current works on implicit semantic expansion derive surrogate upper bounds of these ideal losses and optimize those instead. Implicit semantic expansion methods yield strong results in DG-ReID and are simple to implement, usually consisting of a loss function that plugs naturally into a standard deep neural network architecture such as a ResNet-50~\citep{ResNet}. 

Our previous method, \textbf{D}omain \textbf{E}mbedding E\textbf{X}pansion (DEX)~\citep{DEX}, is one such application of implicit semantic expansion specialized toward DG-ReID, and this work aims to improve on our previously published method by addressing its limitations and increasing its capability. Performing a detailed analysis of DEX, we discover an interesting phenomenon: the DEX loss function causes inter-class distances in the final classifier layer to shrink. Combined with the dynamics of the DEX loss, this shrinkage reduces constraints on the parameters to be learnt, resulting in a complex model with a higher tendency to over-fit. Thus, the theoretical benefits of implicit semantic expansion are offset by this tendency to over-fit and prevent \methodDEX{} from reaching its full potential. Experiments on inter-class weight distances are presented in \autoref{fig:weight-dist-scores-study}, while details of our analysis of the issue are discussed in \autoref{sec:analysis}.

To overcome this limitation, we propose a crucial enhancement: by performing explicit semantic expansion along with the implicit, we can mitigate the shrinkage of inter-class distances during training and extend the runway of improvement during training. 

Our method, \textbf{U}nified \textbf{D}eep \textbf{S}emantic E\textbf{X}pansion (\methodUDSX{}), accommodates both implicit and explicit semantic expansion by proposing three major framework innovations. 

Firstly, \DSDFull{} (\DSDAbrv{}) streams the data into two independent pathways during training, allowing each stream to specialize in implicit or explicit semantic expansion without cross-interference from the other stream. The additional independent explicit semantic expansion stream creates an extra loss function component to mitigate the shrinkage of inter-class distances. 

Secondly, a progressive expansion schedule, \PSTEFull{} (\PSTEAbrv{}), moderates the explicit semantic expansion process, allowing expansion to take place progressively and in targeted feature channels. This stabilizes training during the application of explicit semantic expansion, which if not carefully controlled could degrade performance. 

Finally, a reunification component, \CSRFull{} (\CSRAbrv{}), recombines the two independent semantic expansion streams. \CSRAbrv{} consists of a trio of loss functions, \CSPFull{} (\CSPAbrv{}), \CSCFull{} (\CSCAbrv{}), and \CSTFull{} (\CSTAbrv{}), which enforce stream-invariance of features while maintaining class consistency. Instances from both streams, having gone through different semantic expansion processes, are trained to retain pertinent features that remain invariant to their particular streams. At the same time, these stream-invariant features are encouraged to maintain inter-class alignment so that features within the same class remain aligned with one another. 

With this proposed design, \methodUDSX{} achieves significantly higher test performance in all major DG-ReID benchmarks. Further, we demonstrate that \methodUDSX{} performs well in more general image retrieval benchmarks, significantly surpassing the SOTA in benchmarks such as CUB-200-2011 (CUB)~\citep{WahCUB_200_2011}, Stanford-Cars (Cars196)~\citep{cars196}, VehicleID~\citep{vehicleid} and Stanford Online Products (SOP)~\citep{songCVPR16}.

To summarize, our contributions are as follows:
\begin{enumerate}
\itemsep=0pt
\item We uncover a limitation of DEX, our current state-of-the-art (SOTA) implicit semantic expansion method in DG-ReID. We analyze this limitation, which also applies to other implicit expansion methods, and use our analysis to propose a resolution;
\item We design a dual-stream framework for decoupling and reunifying explicit and implicit semantic expansions. It is simple to implement, addresses the limitations of previous methods and further boosts DG-ReID performance to the next level. To the best of our knowledge, \methodUDSX{} is the first method to unify implicit and explicit semantic expansion in training;
\item Our new method, \methodUDSX{}, outperforms existing SOTA methods in various DG-ReID benchmarks by large margins;
\item \methodUDSX{} also outperforms the SOTA on more general image retrieval tasks such as CUB-200-2011, Stanford Cars, VehicleID and Stanford Online Products for methods with a ResNet-50 backbone architecture;
\end{enumerate}  

\section{Related Work}
\label{sec:related-work}

\subsection{Domain Adaptation Person ReID} Supervised Person ReID methods ~\citep{PCB,StrongBaseline, MGN,OSNet} have shown impressive accuracy in same-domain benchmarks, but these methods overfit to the training domain (dataset) and generalize poorly on different domains. Domain Adaptation Person ReID (DA-ReID) addresses this issue by training models to \textit{adapt} to differences between source and target domains. DA-ReID models are trained on the labeled source domain data and utilize the attributes or styles of the unlabeled target domain data to perform adaptation. Recent DA-ReID methods exploit GAN-based image-synthesis~\citep{SPGAN}, domain alignments~\citep{MMFA,MMFA-AAE}, pseudo labels~\citep{TJ-AIDL} and memory banks~\citep{dai2021IDM,zheng2021Group-awareLabelTransfer,zheng2021ExploitingSampleUncertainty} to close the gap between source and target domains. Although DA-ReID approaches have yielded good performance in recent years, these methods still require a large amount of data from the target domain. It severely affects the model's applicability in the real-world, where access to target domain data is not always possible.

\subsection{Domain Generalization Person ReID} Domain Generalization Person ReID (DG-ReID) aims to learn a model that can perform well in unseen target domains without involving any target domain data for adaptation. DualNorm \citep{DualNorm} first introduced the instance normalization (IN) layer to automatically normalize the style and content variations within the image batch during the training. MMFA-AAE~\citep{MMFA-AAE} used a domain adversarial learning approach to remove domain-specific features. Later DG methods such as DIMN~\citep{DIMN}, QAConv~\citep{QAConv} and M$^3$L~\citep{M3L} used hyper-networks or meta-learning frameworks coupled with memory bank strategies. ACL~\citep{ACL-DGReID} proposed a module to separately process domain invariant and domain specific features, plugging the module to replace selected convolutional blocks. RaMoE~\citep{RaMoE-Dai2021} and META~\citep{META-DGReID} deploy a mixture of experts to specialize to each domain. Style Interleaved Learning (SIL)~\citep{StyleInterleaved} is a framework with a regular forward/backward pass, with one additional forward pass that employs style interleaving on features to update class centroids. Part-Aware Transformer~\citep{ni2023part} learns locally similar features shared across different IDs, further using the part-guided information for self-distillation. Different from the above approaches, which require substantial architectural or framework modifications, applying implicit semantic expansion only requires replacing the cross-entropy loss with a modified loss that encodes the implicit semantic expansion process, without requiring network manipulation or complex training frameworks. Thus, implicit semantic expansion losses can be easily applied to any deep neural network architecture and yield strong results in DG-ReID benchmarks.

\subsection{Implicit Deep Feature Semantic Expansion} Recent works~\citep{DeepAugment,ISDA-TPAMI2020,ISDA-NeurIPS2019} discovered that it was possible to alter the semantics (e.g., color, shape, orientation) of images by shifting their encoded deep features in specific directions. DeepAugment~\citep{DeepAugment} designed an image-to-image model to create novel and semantically meaningful image samples to boost the training set. Implicit semantic data augmentation methods~\citep{ISDA-TPAMI2020,ISDA-NeurIPS2019} go a step further by cutting away the extraneous image-to-image generation and instead generate new and semantically meaningful samples \textit{in feature space}. They rely on class-level deep feature statistics to direct the perturbations in feature space while preserving class label information. \methodDEX{} \citep{DEX} tailored this approach for the DG-ReID problem by semantically expanding the deep features along \textit{domain-wise} directions instead of class-wise directions. We performed a detailed analysis of the implicit semantic expansion loss of the \methodDEX{} method and discovered a limitation of this technique for the DG-ReID problem. Based on our discovery, we proposed \methodUDSX{}: a dual-stream framework that accommodates both explicit and implicit semantic expansion techniques to boost the model's generalization ability. We present our analysis in the following section. 

\section{Background}
\label{sec:dex}
\subsection{Domain Embedding Expansion} 

\begin{figure}[t]
\centering
\includegraphics[trim={18.5cm 0 0 0},clip, width=0.7\linewidth]{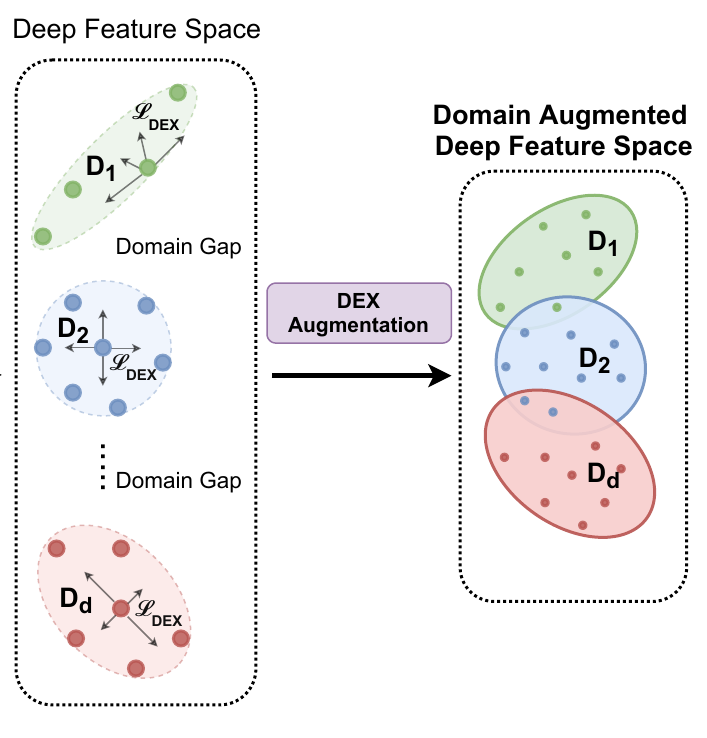}
\caption{\methodDEX{}, our adaptation of implicit semantic expansion to DG-ReID. \methodDEX{} improves exploration of the domain space by implicitly projecting training points in the directions of the domain distribution.}
\label{fig:dex-overall}
\end{figure}

Our previous method, Domain Embedding Expansion (\methodDEX{}), is an implicit deep feature expansion technique that improves cross-domain feature representation learning. It is specially adapted to close domain gaps in multi-domain generalization problems. Like other such methods~\citep{DFI,DeepAugment,ISDA-NeurIPS2019}, the \methodDEX{} loss function implicitly perturbs features in semantically meaningful directions sampled from per-domain feature statistics. \methodDEX{} is a perfect drop-in replacement for the identity classification loss of standard Person ReID baselines and is an efficient way to perform semantic expansion without having to explicitly manipulate feature semantics. Figure \ref{fig:dex-overall} presents a high level overview of \methodDEX{}. 

The \methodDEX{} loss presented in our previous paper is derived as such. Given a feature extractor $f$ with classifier layer weights $\mathbf{w}$, consider the cross-entropy loss of a single sample $\mathbf{x}$ with label $y$:
\begin{equation} 
\Lagr_{CE}= -\log \left( \frac{e^{\mathbf{w}_{y} \cdot f(\mathbf{x}) }}{\sum_{j=1}^{C} e^{\mathbf{w}_{j} \cdot f(\mathbf{x})}} \right),
\label{eqn:softmax}
\end{equation}
where $C$ is the number of unique person identities indexed by $j$. Biases are omitted in our classifier layer. The design rationale behind \methodDEX{} is derived from \citep{ISDA-NeurIPS2019}. Given that $\mathbf{x}$ comes from domain $d$, we consider the \emph{expected} softmax loss if the deep features were projected along domain-level covariance directions $\Sigma_{d}$: 
\begin{equation} 
\begin{split}
\Lagr_{\infty} & = \mathbb{E}_{ \widetilde{f(\mathbf{x})}  } \left[ -\log \left( \frac{e^{\mathbf{w}_{y} \cdot 
\widetilde{f(\mathbf{x})} }}{\sum_{j=1}^{C} e^{\mathbf{w}_{j} \cdot \widetilde{f(\mathbf{x})}  }} \right) \right] \\
& = \mathbb{E}_{ \widetilde{f(\mathbf{x})}}  \left[ \log \left(  
\sum_{j=1}^{C} e^{ (\mathbf{w}_{j} - \mathbf{w}_{y}) \cdot \widetilde{f(\mathbf{x})} } \right) \right],
\end{split}
\label{eqn:loss-infinity}
\end{equation}
where $ \widetilde{f(\mathbf{x})} \sim \mathcal{N}( f(\mathbf{x}) , \lambda \Sigma_{d}) $ are the projected features assumed to be normally distributed around $f(\mathbf{x})$ with domain-conditional covariance $\Sigma_{d}$, and $\lambda \ge 0$ controls the magnitude of the projection. Applying Jensen's Inequality, $\mathbb{E}[\log (X)] \le \log (\mathbb{E} [X])$, we can move the logarithm out of the expectation to get:

\begin{equation} 
\Lagr_{\infty} \le \log \left( \sum_{j=1}^{C} \mathbb{E}_{ \widetilde{f(\mathbf{x})}  } e^{ (\mathbf{w}_{j} - \mathbf{w}_{y}) \cdot \widetilde{f(\mathbf{x})} }  \right)
\label{eqn:loss-jensen}
\end{equation}

We apply the moment generating function $\mathbb{E}[\exp(tX)]=\exp(t \mu + \frac{1}{2} \sigma^2 t^2)$, $X \sim \mathcal{N}(\mu,\sigma^2)$, substituting $t$ with $(\mathbf{w}_{j} - \mathbf{w}_{y})$ and $X \sim \mathcal{N}(\mu, \sigma^2)$ with $ \widetilde{f(\mathbf{x})} \sim \mathcal{N}( f(\mathbf{x}) , \lambda \Sigma_{d})$, to derive our loss: 
\begin{multline}
\Lagr_{\methodDEX{}} = 
-\log \Bigl( \frac{e^{\mathbf{w}_{y} \cdot f(\mathbf{x}) }}{\sum\limits_{j=1}^{C} e^{\mathbf{w}_{j} \cdot f(\mathbf{x}) + \frac{\lambda}{2} (\mathbf{w}_{j}^\intercal-\mathbf{w}_{y}^\intercal) \Sigma_{d} (\mathbf{w}_{j}-\mathbf{w}_{y}) }} \Bigr) 
\label{eqn:dex}
\end{multline}
$\Lagr_{\methodDEX{}}$ is an upper bound to the \emph{expectation} of cross-entropy loss over infinitely perturbing $f(\mathbf{ x })$ in directions determined by per-domain covariance matrix $\Sigma_{d}$. In theory, optimizing $\Lagr_{\methodDEX{}}$ reaps the benefits of semantic data expansion while avoiding the overheads of having to perform the computation explicitly.

\subsection{\methodDEX{} Loss Function Limitations}
\label{sec:analysis}

We observed a unique behavior of models trained with the \methodDEX{} loss. While achieving strong generalization performance, the best validation performance of these models is achieved very quickly at lower epochs (Table~\ref{tab:lambda-comparison}), showing that it over-fits to the source datasets at an early stage. This phenomenon occurs consistently in all benchmarks we tested. In contrast, if we replace the \methodDEX{} loss with a baseline person identity cross-entropy loss, the performance improves for much longer before over-fitting, even though it ultimately reaches a lower performance. A question naturally arises from this observation: is it possible to circumvent this limitation of the \methodDEX{} loss and allow it to continue improving beyond this limit? In this follow-up study, we analyze the \methodDEX{} loss and formulate a hypothesis about the cause behind this behavior. To rectify this, we design a method that unifies both implicit and explicit semantic expansion processes and allows the model to further improve in generalization. 

\begin{figure}[ht]
    \centering
    \textbf{Inter-Class Distances Over \methodDEX{} Strength}\par
    \includegraphics[scale=0.5]{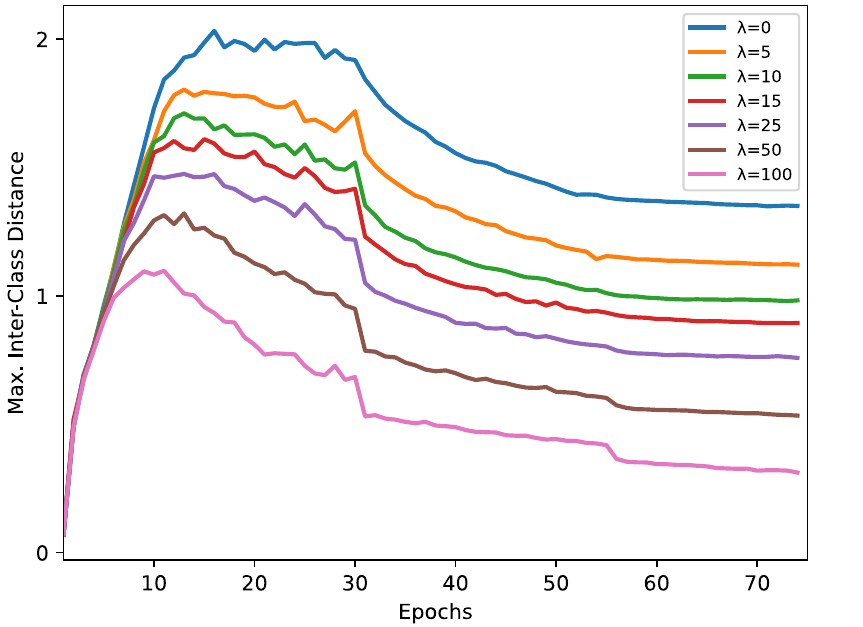}
    \label{subfig:weight-dist-plot}
    \caption {Increasing DEX strength decreases the distance between class weights. Best viewed in color.}
    \label{fig:weight-dist-scores-study}
\end{figure}

For the purposes of this analysis, we can simplify the \methodDEX{} loss from Equation~\ref{eqn:dex} as such:
\begin{equation}
\Lagr_{\methodDEX{}} = -\log \biggl( \frac{e^{\mathbf{w}_y \cdot f(\mathbf{x}) }} { \sum\limits_{j=1}^{C} e^{\mathbf{w}_j \cdot f(\mathbf{x}) + \mathbf{\Gamma}} } \biggl)
\label{eqn:dex-simple}
\end{equation}
Where $\mathbf{\Gamma}=\frac{\lambda}{2}(\mathbf{w}_{j}^\intercal-\mathbf{w}_{y}^\intercal) \Sigma_{d} (\mathbf{w}_{j}-\mathbf{w}_{y})$. Recast in this form, we can see that the \methodDEX{} loss is simply the original cross-entropy (CE) loss for identity classification with an extra term added for each value in the denominator sum. The first crucial thing to observe is that the covariance matrix $\Sigma_d$ is positive semi-definite, which implies that $\mathbf{\Gamma} \ge 0$, since $\lambda$ is non-negative. Similar to CE loss, the \methodDEX{} loss maximizes the similarity between feature and true-class weights in the numerator and minimizes the denominator. But in order to minimize the denominator, Equation~\ref{eqn:dex-simple} has to minimize $\mathbf{\Gamma}$. To do this, it has to minimize the difference between $w_y$ and $w_j$, which encourages class weights to be close to each other. Consider the scenario where all class weight vectors $w_j$ have converged very closely to one another. Given an input $\mathbf{x}$ of label $y$, the backbone extractor $f$ would have to exaggerate the magnitude of $f(\mathbf{x})$ in order for $\mathbf{w}_y \cdot f(\mathbf{x})$ in the numerator to be greater than $\mathbf{w}_{j \neq y} \cdot f(\mathbf{x})$ in the denominator. $f$ is encouraged to learn weights of larger magnitude, undoing the weight-decay regularization that tries to constrain model parameters, and we are left with a complex backbone model that over-fits early as an unintended side effect of optimizing the \methodDEX{} loss. \autoref{tab:lambda-comparison} shows that applying \methodDEX{} is effective, but too much of it rapidly degrades performance, and that the scores saturate faster as the \methodDEX{} loss strength increases. \autoref{fig:weight-dist-scores-study} plots and compares the maximum inter-class weight distances over time for \methodDEX{} as we vary its strength. As we increase the strength of \methodDEX{} through the hyperparameter $\lambda$, we observe the inter-class weight distances shrinking.

\begin{table}[t]
    \centering
    \caption{Best epochs and scores on C+D+MS $\rightarrow$ M over values of $\lambda$ (see Sec~\ref{sec:experiment-settings} for benchmark details)}
    \begin{adjustbox}{width=1\linewidth}
    \begin{tabular}{c|c|c|c}
        \hline
        $\lambda$     & Best Epoch & mAP & Rank-1 \\
        \hline
        No \methodDEX{} ($\lambda=0$)     & 57 & 54.0 & 79.8 \\
        \textbf{5}     & 39 & \textbf{54.8} & \textbf{81.1} \\
        15 & 39 & 54.2 & 80.2 \\
        25     & 35 & 54.1 & 79.7 \\        
        50 & 34 & 52.7 & 78.1 \\ 
    \hline
    \end{tabular}
    \end{adjustbox}
    \label{tab:lambda-comparison}
\end{table}

\section{Integrating Explicit Semantic Expansion}
A way to alleviate the issue described in the previous section would be to reduce the hyperparameter $\lambda$. However, \autoref{tab:lambda-comparison} shows that setting $\lambda$ too low also hurts performance; the extra $\mathbf{\Gamma}$ term improves results albeit with the cost of early over-fitting. Implicit semantic expansion methods were formulated as upper bounds of an idealized loss, and the deviation between the upper bound and the true loss ($L_{\infty}$) is one of the main contributors of this limitation. We propose to add explicit semantic feature expansion to restore freedom to the inter-class weights and reach a closer approximation to the idealized loss. From the robustness literature, it is well known that injecting features with noise during training can improve model generalization~\citep{he2019ParametricNoiseInjection,wu2020adversarial}. Building on this, we design an explicit semantic expansion method that computes feature statistics at the domain level and uses these statistics to apply the explicit semantic expansion on features in a targeted way. 

\begin{figure}[ht]
  \centering
   \includegraphics[width=1\linewidth]{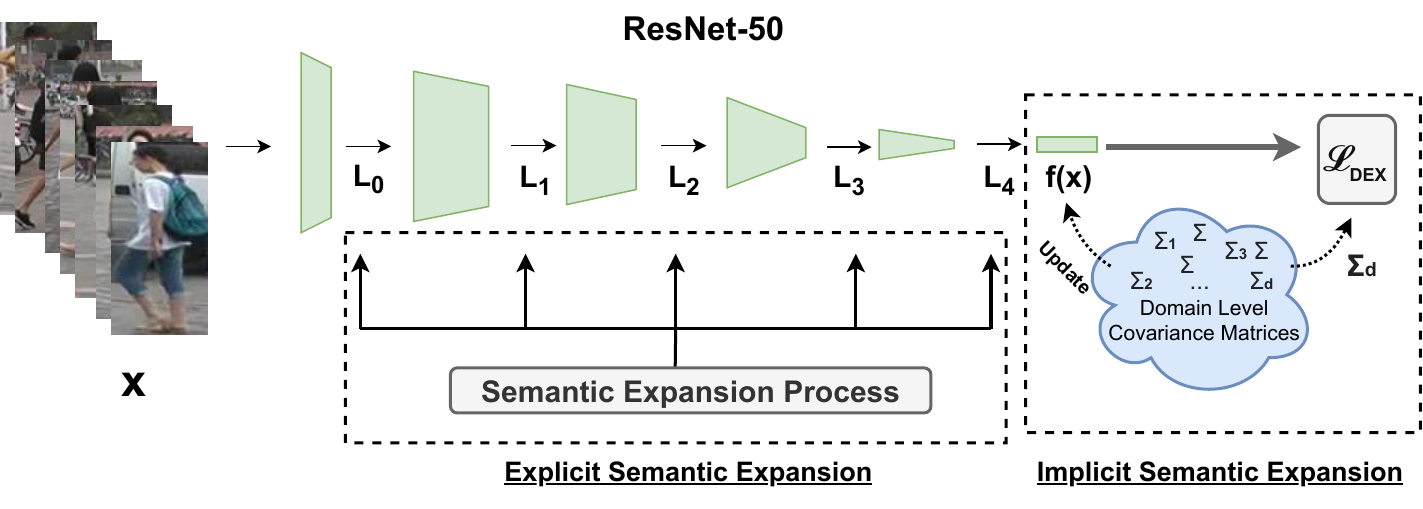}
   \caption{Naively combining explicit and implicit semantic expansion yields poor results as they disrupt each other.}
   \label{fig:naive-onecol}
\end{figure}

\begin{figure*}[ht]
  \centering
   \includegraphics[width=1\linewidth]{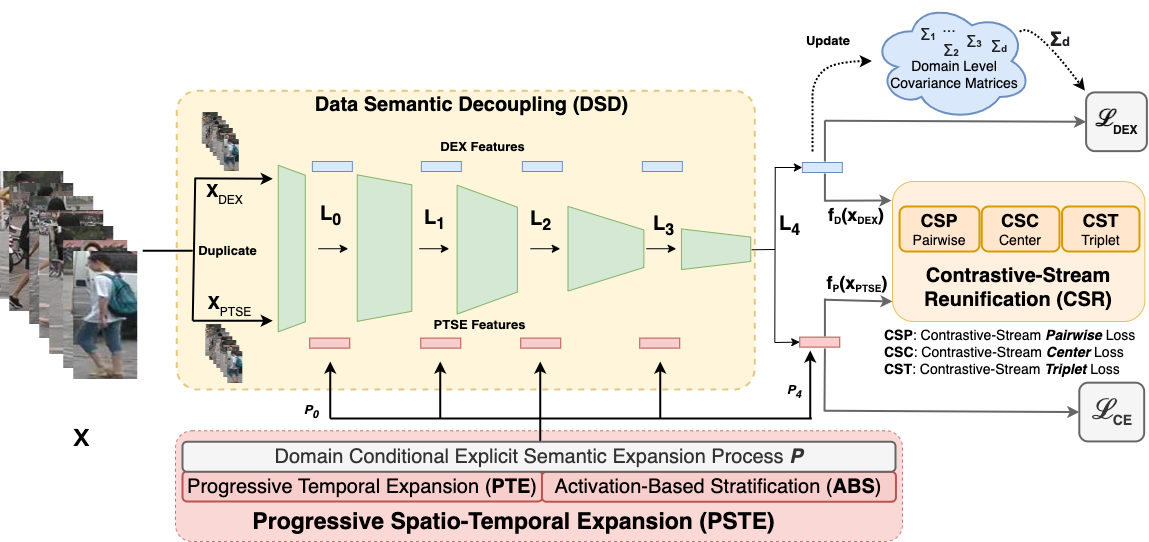}
    \caption{Our method \methodUDSX{} unifies implicit and explicit semantic expansion in 3 parts: (1) \DSDFull{} (\DSDAbrv{}) isolates Domain Embedding Expansion (\methodDEX{}) and \PSTEFull{} (\PSTEAbrv{}) into dedicated streams, (2) \PSTEAbrv{} is the engine behind the explicit semantic expansion, and (3) \CSRFull{} (\CSRAbrv{}) reunifies the streams at the end and more details on this component are illustrated in \autoref{fig:contrastive-stream-reunification-overview}. Best viewed in color.}
   \label{fig:udsx-twocol}
\end{figure*}

A naive way to combine both implicit and explicit semantic expansion would be to perform both in conjunction as the model trains, as shown in \autoref{fig:naive-onecol}. However, we found that this did not improve performance and was detrimental in some cases, as \autoref{tab:ablation-components} shows. By prematurely expanding the semantics of the intermediate features, we alter the distribution of the resultant feature embeddings that are fed to the \methodDEX{} loss, thereby disrupting the learning of the implicit semantic loss. This over-coupling of implicit and explicit expansion processes constrains the model and renders it incapable of fully benefiting from both types of expansion. Instead, our proposed \methodUDSXFull{} (\methodUDSX{}) synergizes the strengths of both types of semantic expansion and enables the model to train for extended periods and improve past previous limits. \methodUDSX{} comprises three framework innovations: 

\begin{itemize}
\item \DSDFull{} (\DSDAbrv{})
\item \PSTEFull{} (\PSTEAbrv{})
\item \CSRFull{} (\CSRAbrv{})
\end{itemize}

\subsection{\DSDFull{}}
Observing that explicit semantic expansion interferes with \methodDEX{}, we designed a framework to better accommodate both types of expansion so that they can work together. Given a batch of data, we duplicate its contents into two streams and process them separately such that each stream receives the same content while having the flexibility to apply independent training conditions. Our twin data stream framework, \DSDFull{} (\DSDAbrv{}), is shown in \autoref{fig:udsx-twocol}. With this modification, the \methodDEX{} stream trains with \methodDEX{} loss without interference from the explicit stream, and vice versa. The backbone model's weights are shared between both data streams during training: during inference, \DSDAbrv{} is disabled as only one stream is required to produce the output features.
\subsection{\PSTEFull{}}
\label{subsec:method-semantic-expansion-policy}
We denote our proposed explicit semantic expansion process by the symbol $P$: given a feature $\mathbf{x}$, $P(\mathbf{x})$ represents the output of applying our explicit semantic expansion on $\mathbf{x}$. More details are provided in the following sub-sections. In \autoref{subsubsec:explicit-semantic-expansion}, we describe the process of generating perturbations conditioned on the domain of the input and adding them to intermediate features. Next, in \autoref{subsubsec:progressive-stochastic-semantic-expansion}, we describe the progression of semantic expansion across intermediate model layers over the course of training using our \PTEFull{} (\PTEAbrv{}). Finally, we describe our \ABSFull{} (\ABSAbrv{}) in \autoref{subsubsec:activation-based-stratification}, which applies explicit semantic expansion on intermediate tensors in a controlled and localized manner that harmonizes with the backbone network's learning dynamics. 
\subsubsection{Domain Conditional Explicit Semantic Expansion Process}
\label{subsubsec:explicit-semantic-expansion}
The intermediate features of the backbone are the targets of the explicit semantic expansion. We dynamically compute feature statistics of intermediate layers of the model by tracking and updating on-the-fly the per-element variances of each intermediate feature: $V^d_{k \in \{0,1,2,3,4\}}$, with $k$ indexing the intermediate layers of the backbone ResNet-50 and $d$ indexing the domain of the sample. We sample a zero-mean Gaussian using these variances conditioned on domain $d$, obtaining semantically meaningful perturbations which are added in a controlled manner to the intermediate features at layer-$k$. 

\subsubsection{\PTEFull{} (\PTEAbrv{})}
\label{subsubsec:progressive-stochastic-semantic-expansion}
We target the semantic expansion to be applied progressively on lower to higher layers over time. This design is guided by the observation that models first learn to encode low-level concepts such as textures, edges, shapes and colors in early layers before combining them into higher-order semantic abstractions in later layers. We harmonize with this natural order of development, randomly selecting from lower layers to expand at the start of training and progressively including higher layers over time. To moderate the amount of semantic change at any time, our policy randomly selects only one layer to perform semantic expansion at each forward pass.

\subsubsection{\ABSFull{} (\ABSAbrv{})}
\label{subsubsec:activation-based-stratification}
In order to maintain a controlled level of semantic change, we perturb only selected channels, or strata, of the intermediate tensor based on their average activation values. The stratification is specified for each intermediate layer as they each have different numbers of channels. Early layers have fewer channels compared to later layers, so a narrower band of channels would be selected for semantic expansion. Channels are sorted and selected according to their average activation values and selected only if these values fall within a defined quantile. Our stratified design allows for flexibility, allowing us to apply semantic expansion in a targeted way. In our specific case we select the middle quantile of channels whose average activation values fall within the top [$\frac{3}{8}C_k$,$\frac{5}{8}C_k$], where $C_k$ is the number of channels in the $k$-th intermediate tensor. In other words, we skip the channels with average activation values in the top and bottom quartiles. The intuition behind this design is that the channels with the highest average activation values could contain important semantic information and should not be disrupted, while the lowest activated channels are unlikely to contain meaningful semantic information for expansion.

\subsubsection{Overview of Components}
\begin{figure*}[ht]
  \centering
   \includegraphics[width=1\linewidth]{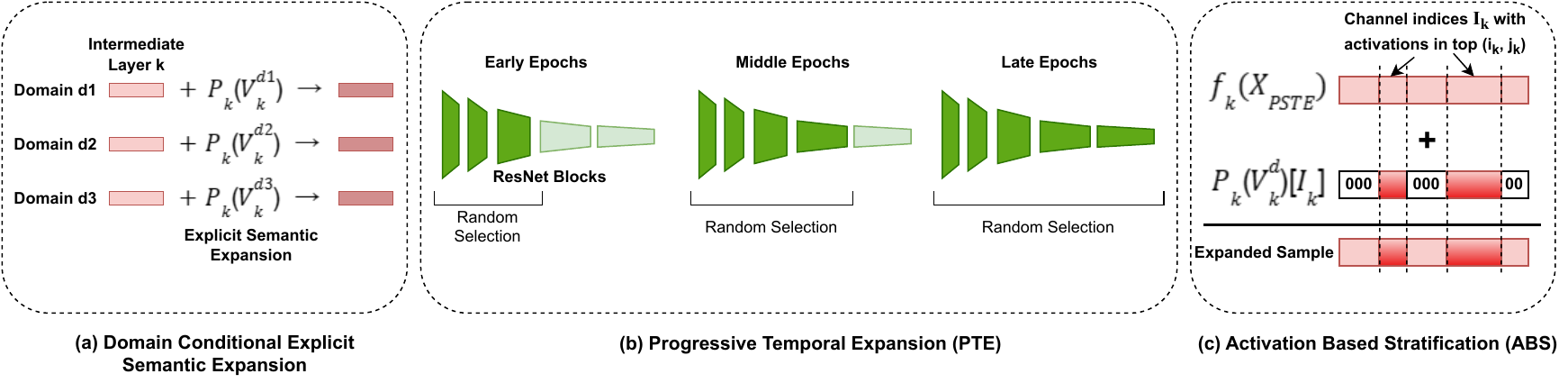}
    \caption{Components of \PSTEAbrv{} illustrated. (a) Intermediate features are perturbed based on the domain that they come from, allowing the semantics of each domain to be better explored over time. (b) \PTEAbrv{}: We start by randomly selecting among early intermediate features to perturb, and gradually over time expand the selection set to the later layers as the model begins to understand higher-level semantics. (c) \ABSAbrv{}: Only selected channels are perturbed, based on their activation values falling within a pre-defined quantile: channels that are either very important or meaningless are left unperturbed. For more details, refer to Section~\ref{subsec:method-semantic-expansion-policy} and Algorithm~\ref{alg:semantic-expansion-policy} }
   \label{fig:pste-illustration}
\end{figure*}

\begin{algorithm}
\caption{\PSTEFull{} (\PSTEAbrv{}). In all our experiments, we fix $R$ to be a uniform random selector. The channel activation sorter $\Lambda$ abstracts the process of ranking the channels of an intermediate tensor $x_k$ by their average activation values and returning the indices to the $i_k$-th to $j_k$-th largest channels (for $i_k < j_k$). \\
\textbf{Abbreviations}: \\ SE=Semantic Expansion (See~\ref{subsubsec:explicit-semantic-expansion}) \\
\PTEAbrv{}=\PTEFull{} (See~\ref{subsubsec:progressive-stochastic-semantic-expansion}) \\
\ABSAbrv{}=\ABSFull{} (See~\ref{subsubsec:activation-based-stratification})}
\label{alg:semantic-expansion-policy}
\textbf{Definitions} \\
$\mathbf{x}$: Input image, \\
$d$: Domain of $\mathbf{x}$, \\
$f$: Backbone model, \\
$V$: Per-element variances, \\
$\mathcal{N}$: Gaussian model, \\
$L$: Ordered list of layer indices, \\
$\mu$: Minimum layer width, \\
$t$: Current epoch, \\
$T$: Num epochs for \PSTEAbrv{}, \\
$R$: Random selector \\
$S$: Top-k strata for each intermediate layer \\
$\Lambda$: Average channel activation sorter \\
\begin{algorithmic}[1]
\STATE \COMMENT{\textbf{PTE}: select one layer from candidate layers based on stage of training.} \\
\STATE $l \gets (|$L$| - 1) \times min(\lfloor \frac{t}{T} \rfloor, 1)$ \COMMENT{Get last index}
\STATE $l \gets max(l,\mu)$ \COMMENT{Ensure min selection width}
\STATE $k \gets$ R(L[$0,...,l$]) \COMMENT{Choose one layer only}
\STATE \COMMENT{\textbf{ABS}: apply SE only on pre-defined target channels based on per-channel activation values.} \\
\STATE $(i_k, j_k) \gets S(k)$ \COMMENT{Strata for layer-k, $i_k$ $\leq$ $j_k$}
\STATE $\mathbf{x}_k \gets f_{:k}(\mathbf{x})$ \COMMENT{Intermediate output at k}
\STATE $I_k \gets \Lambda(\mathbf{x}_k,i_k,j_k)$ \COMMENT{Top-$i_k$-to-$j_k$ channel indices}
\STATE $\mathbf{x}_k$[$I_k$] $\gets \mathbf{x}_k$[$I_k$] + $\mathcal{N}(0, \sqrt{V^d_k})$[$I_k$] \\
\RETURN
\end{algorithmic}
\end{algorithm}

Figure~\ref{fig:pste-illustration} illustrates the core ideas behind \PSTEAbrv{}. In (a), we convey the idea that each intermediate feature is explicitly expanded using a domain-conditional perturbation as described in~\autoref{subsubsec:explicit-semantic-expansion}. In (b), we illustrate the progressive and temporal nature of \PTEAbrv{}, where the perturbations are only randomly applied in early layers at the start of training and expand to the later layers as the model learns to encode higher-order semantics. In (c), we illustrate how \ABSAbrv{} is used to select channels for semantic expansion based on their activation values.

In more detail, Algorithm~\ref{alg:semantic-expansion-policy} describes how \PSTEAbrv{} is applied to a sample input during a forward pass. Having randomly selected an intermediate layer $k$ to perform the expansion, given an input $\mathbf{x}$ from domain $d$ we intercept the intermediate feature $\mathbf{x}_k$ at layer-$k$ and rank their channels according to their average activation values to get the top $i_k$-to-$j_k$ highest channel indices $I_k = \Lambda(\mathbf{x_k}, i_k, j_k)$. Using the stored variances $\Sigma_{d,k}$, we sample an expansion direction and add it to $\mathbf{x_k}$ only at the desired channel indices, finally deriving the expanded sample $P(\mathbf{x}_k) = \mathbf{x}_k + \mathcal{N}(0,\Sigma_{d,k})_{I_k}$.

\section{\CSRFull{}}
\begin{figure*}[ht]
    \centering
    \includegraphics[width=1\linewidth]{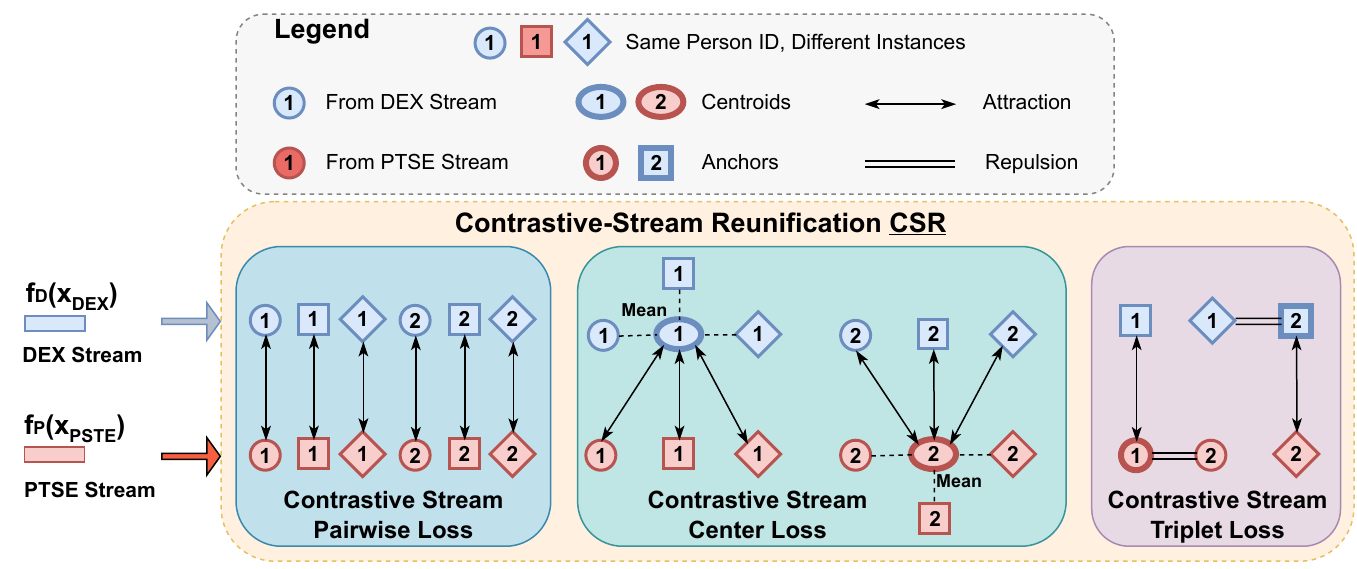}
    \caption {Overview of our \CSRFull{} (\CSRAbrv{}) Loss.}
    \label{fig:contrastive-stream-reunification-overview}
\end{figure*}

After the passing through the backbone model, features are compared across \methodDEX{} and \PSTEAbrv{} streams using our \CSRFull{} (\CSRAbrv{}) module to teach the model to reconcile cross-stream differences and extract key relevant information. \CSRAbrv{} is a combination of three losses: Contrastive-Stream Pairwise (CSP) Loss, Contrastive-Stream Center (CSC) Loss and Contrastive-Stream Triplet (CST) Loss, all of which operate across \methodDEX{} and \PSTEAbrv{} streams to achieve this goal. CSP encourages the learning of expansion-invariant features, while CSC and CST create a more challenging metric learning environment for the model to learn similarities between corresponding samples across streams. \autoref{fig:contrastive-stream-reunification-overview} sketches an overview of the components of \CSRAbrv{} Loss.

\subsection{\CSPFull{}}
In order to encourage the model to resolve the differences between features derived from the \methodDEX{} and \PSTEAbrv{} streams, we apply a $L_1$ distance loss between pairs of duplicated samples that have each passed through different streams. Given $f_D(\mathbf{x})$ and $f_P(\mathbf{x})$ are global features from the implicit and explicit streams of the base model $f$ respectively, our \textbf{\CSPFull{} (\CSPAbrv{})} is defined as:
\begin{equation}
    L_{\CSPAbrv{}}(f,\mathbf{x}) = \lVert f_D(\mathbf{x}) - f_P(\mathbf{x})\rVert_1
    \label{eq:feat-dist-loss}
\end{equation} 
\subsection{\CSCFull{}}
Next, we apply a modified center loss, our \textbf{\CSCFull{}} (\CSCAbrv{}), to encourage the model to learn expansion-invariant feature representations. Given a batch of $m$ samples $\mathbf{x}$ with labels $\mathbf{y}$ and batch-wise class centroids $\mathbf{c^{y_i}} = \frac{1}{m} \sum\limits_{i=1}^{m} f(\mathbf{x}_i)$, the original center loss~\citep{Wen2016ARecognition} has the following form: 
\begin{equation}
    L_{Cen}(f,\mathbf{x},\mathbf{y}) =\sum\limits_{i=1}^{m} \, \norm{\, f(\mathbf{x}_i) - \mathbf{ c^{y_i}} \, }_2
    \label{eq:center-loss}
\end{equation}
In \CSCAbrv{}, our loss is the distance between each sample $\mathbf{x}$ and the class centroid from its \textit{opposing} stream:
\begin{align}
    L_{\CSCAbrv{}}(f,\mathbf{x},\mathbf{y}) = \sum\limits_{i=1}^{m} \, \norm{\, f_D(\mathbf{x}_i) - \mathbf{ c^{y_i}_{P}} \, }_2 + \\ 
    \sum\limits_{i=1}^{m} \, \norm{\, f_P(\mathbf{x}_i) - \mathbf{ c^{y_i}_{D}} \, }_2
    \label{eq:center-loss-mean}
\end{align}
where $\mathbf{c^{y_i}_D} = \frac{1}{m} \sum\limits_{i=1}^{m} f_D(\mathbf{x}_i)$, $\mathbf{c^{y_i}_P} = \frac{1}{m} \sum\limits_{i=1}^{m} f_P(\mathbf{x}_i)$ are the centroids from the \methodDEX{} and \PSTEAbrv{} streams respectively.
\subsection{\CSTFull{}}
Finally, we apply our own variant of the triplet loss~\citep{TriNet}. Given a feature extractor $f$, a distance metric $\delta$, and a triplet $\mathbf{t} = (a,p,n)$ where $a$ and $p$ are from the same class and $n$ is from a different class, $[\,\delta_{f(a),f(p)} \, - \, \delta_{f(a),f(n)}\,]_+$ computes the triplet loss. Our \textbf{\CSTFull{} (\CSTAbrv{})} loss is designed to optimize similarity in corresponding features from opposing streams. Given two feature extractors $f$ and $g$, we define the cross-extractor triplet loss $L_{XT}$ as:
\begin{equation}
    L_{XT}(f,g,d,\mathbf{t}) = [\,\delta_{f(a),g(p)} - \delta_{f(a),f(n)}\,]_+
    \label{eq:dual-extractor-loss}
\end{equation}
We then define the \CSTFull{} as follows:
\begin{multline}
L_{\CSTAbrv{}}(f, \mathbf{x}, \mathbf{y}, \delta, \mathbf{T})  = \\  
\frac{1}{2} \: \mathbb{E}_{\mathbf{t} \in \mathbf{T(x,y)}}  [  L_{XT}(f_D,f_P,\delta,\mathbf{t}) + 
L_{XT}(f_P,f_D,\delta,\mathbf{t}) ]
\label{eq:cross-stream-trip-loss}
\end{multline}
where $\mathbf{T}(\mathbf{x},\mathbf{y})$ is a set of triplets we obtain by applying Batch-Hard example mining~\cite{TriNet}. 

In forcing anchor and positive samples to be from opposite streams while anchor and negative are taken from the same stream, our \CSTFull{} increases the difficulty of the triplet optimization task and further encourages the features to be invariant to both explicit and implicit semantic expansions. 

With non-negative weights $\psi_1, \psi_2, \psi_3$ to control the strength of each loss component, the \textbf{Cross-Stream Reunification (\CSRAbrv{})} loss is:
\begin{equation}
L_{CSR} = \psi_1 \hyperref[eq:feat-dist-loss]{L_{\CSPAbrv{}}} + \psi_2 \hyperref[eq:center-loss-mean]{L_{\CSCAbrv{}}} + \psi_3 \hyperref[eq:cross-stream-trip-loss]{L_{\CSTAbrv{}}}
\label{eq:unification-loss}
\end{equation}

\section{Combined Loss Function} 
Our combined loss consists of a semantic expansion loss term and a stream re-unification loss term. For the semantic expansion loss, we combine the explicit and implicit semantic expansion losses into a single loss term we call the Semantic Expansion (SE) loss:
\begin{equation}
L_{SE}(\mathbf{x}) = \beta_1 \hyperref[eqn:softmax]{L_{CE}}(f_P(\mathbf{x})) + \beta_2 \hyperref[eqn:dex-simple]{L_{\methodDEX{}}} (f_D(\mathbf{x})),
  \label{eq:nd-loss}
\end{equation}
where $f_D$ and $f_P$ represent forward passes of data samples over the backbone, but isolated only within the implicit (\methodDEX{}) and explicit (\PSTEAbrv{}) streams respectively. $\beta_1, \beta_2 \ge 0$ balance the mix of implicit and explicit semantic expansion. The presence of the CE loss applied on the \PSTEAbrv{} stream acts as a counterbalance, preventing the class weight vectors from converging too closely together due to the \methodDEX{} loss. Therefore, it helps in stabilizing the overall optimization objective, preventing the model from over-fitting too quickly, and restoring the representation power of the model. The hyperparameter values and relevant configurations for each stream are discussed in \autoref{sec:implementation-details}.

\textbf{Overall Loss Function.} With the hyperparameters $\beta_1, \beta_2, \psi_1, \psi_2, \psi_3$ controlling the relative strengths of each component loss under the hood, the overall loss function of \methodUDSX{} is presented as: 
\\
\begin{center}
$L_{\methodUDSX{}} = \hyperref[eq:nd-loss]{L_{SE}} + \hyperref[eq:unification-loss]{L_{\CSRAbrv{}}}$
\end{center}

To assist in keeping track of all the components and abbreviations introduced in this work, \autoref{tab:abrv-full-desc} presents a list of abbreviations and their full form along with a short and informative description of their intended function in our proposed \methodUDSX{} framework. The reader is also encouraged to look at~\autoref{fig:udsx-twocol} as it graphically presents the relationships between all our components.

\begin{table*}[ht]
\caption{Abbreviation list of important components with short descriptions.}
\begin{tabular}{c|l|p{0.6\linewidth}}
\hline
Abbrv. & \multicolumn{1}{c|}{Full Form}        & \multicolumn{1}{c}{Description} \\ \hline
\methodDEX{}    & Domain Embedding Expansion            & Our existing implicit semantic expansion method~\citep{DEX}                                \\ \hline
\DSDAbrv{}    & Data Semantic Decoupling              & Framework enhancement that manages implicit/explicit semantic expansion in independent streams                                \\ \hline
\PSTEAbrv{}   & \PSTEFull{} & Explicit semantic expansion applied in a controlled and localized manner. Consists of PSE and ABS                                \\
\PTEAbrv{}    & \PTEFull{}      & Explicit semantic expansion applied on lower layers at start, progressing to higher layers over time                                \\
\ABSAbrv{}    & \ABSFull{}       & Explicit semantic expansion applied only to selected strata of intermediate features, based on activation value                                \\ \hline
\CSRAbrv{}    & \CSRFull{}      & Overarching component for reunifying sibling features from implicit and explicit streams. Consists of \CSPAbrv{}, \CSCAbrv{} and \CSTAbrv{}                               \\
\CSPAbrv{}    & \CSPFull{}    & Feature loss ($L_1$) between sibling pairs of features                                \\
\CSCAbrv{}    & \CSCFull{}      & Modified Center Loss~\citep{Wen2016ARecognition} attracting features of one stream to the centroids the other stream, and vice versa                                \\
\CSTAbrv{}    & \CSTFull{}     & Modified Triplet Loss~\citep{TriNet} where anchor and positive are selected from opposing streams, while anchor and negative are from the same stream  
\\ \hline                             
\end{tabular}
\label{tab:abrv-full-desc}
\end{table*}

\section{Experiments}
\subsection{Evaluation Benchmarks}
\label{sec:experiment-settings}
\subsubsection{DG-ReID Datasets and Evaluation}
We perform two major benchmark evaluations for DG-ReID. The traditional evaluation is conducted by training models on the combined datasets of Market-1501~\citep{Market-1501}, DukeMTMC-reID~\citep{DukeMTMC-reID}, CUHK02~\citep{CUHK02}, CUHK03~\citep{CUHK03} and CUHK-SYSU~\citep{CUHK-SYSU}, for a total of 121,765 images covering 18,530 person IDs. The train and test data are combined for all source datasets. The trained model is then tested on four benchmarks i-LIDS~\citep{i-LIDS}, PRID~\citep{PRID}, QMUL-GRID~\citep{GRID} and VIPeR~\citep{VIPeR}. Due to their relatively small sizes, the test sets are shuffled in ten randomized trials and we report the mean.

M$^3$L~\citep{M3L} broke from the traditional evaluation and used large-scale datasets for testing to remove uncertainty from randomized trials. In their work, the model was trained on three out of four large datasets (Market-1501, DukeMTMC-reID, CUHK03 and MSMT17$\_$V2~\citep{MSMT17}, which we abbreviate to M, D, C and MS respectively) and tested the model on the dataset that was left out during training. For example, C+D+MS indicates that the model was trained on a combination of CUHK03, DukeMTMC-reID and MSMT17$\_$V2, and therefore should be tested independently on the Market-1501 dataset. In this evaluation, only the training splits are used for training while the query/gallery split of the test dataset is used for evaluation. Four evaluations are performed, with each dataset taking turns as the target test set. Statistics of four datasets are reported in \autoref{tab:stats-modern-dg-reid}. 
\begin{table}[ht]
\caption{Statistics for DG-ReID datasets.}
\resizebox{\columnwidth}{!}{
\begin{tabular}{c|cc|cc|cc}
\hline
\multicolumn{1}{c|}{\multirow{2}{*}{Dataset}} & \multicolumn{2}{c|}{Train}        & \multicolumn{2}{c|}{Query}        & \multicolumn{2}{c}{Gallery}       \\ \cline{2-7} 
\multicolumn{1}{c|}{}                         & \multicolumn{1}{c|}{ID}   & Image & \multicolumn{1}{c|}{ID}   & Image & \multicolumn{1}{c|}{ID}   & Image \\ \hline
C                                           & \multicolumn{1}{c|}{1367} & 13132 & \multicolumn{1}{c|}{700}  & 1400  & \multicolumn{1}{c|}{700}  & 5332  \\
D                                           & \multicolumn{1}{c|}{702}  & 16522 & \multicolumn{1}{c|}{702}  & 2228  & \multicolumn{1}{c|}{1110} & 17661 \\
M                                           & \multicolumn{1}{c|}{751}  & 12936 & \multicolumn{1}{c|}{750}  & 3368  & \multicolumn{1}{c|}{750}  & 15913 \\
MS                                          & \multicolumn{1}{c|}{1041} & 32621 & \multicolumn{1}{c|}{3060} & 11659 & \multicolumn{1}{c|}{3060} & 82161 \\ \hline
\end{tabular}
}
\label{tab:stats-modern-dg-reid}
\end{table}

\subsubsection{Image Retrieval Datasets and Evaluation} We selected a diverse range of image retrieval benchmark types to test \methodUDSX{}. CUB-200-2011 (CUB)~\citep{WahCUB_200_2011} consists of natural images of birds, Stanford Cars (Cars196)~\citep{cars196} and VehicleID~\citep{vehicleid} consist of images of cars/vehicles. Stanford Online Products~\citep{songCVPR16} are images of online shopping items. Statistics for these datasets are presented in \autoref{tab:stats-image-retrieval}. In all datasets, training and testing classes are mutually exclusive. VehicleID provides 3 test sets, Small/Medium/Large, that contain 6493/13377/19777 images capturing 800/1600/2400 unique classes respectively. Given a test set, we randomly select an image from each class to form the probe set while the remaining images make up the gallery set, finally reporting the average Rank-1 out of 10 randomized trials. For the remaining three datasets, we follow widely used evaluation methodology~\citep{songCVPR16}.

\begin{table}[ht]
\caption{Statistics for image retrieval benchmarks.}
\resizebox{\columnwidth}{!}{
\begin{tabular}{l|c|cc}
\hline
\multirow{2}{*}{Dataset} & \multirow{2}{*}{Classes} & \multicolumn{2}{c}{Images}           \\ \cline{3-4} 
                         &                          & \multicolumn{1}{c|}{Train}  & Test   \\ \hline
CUB-200-2011             & 200                      & \multicolumn{1}{c|}{5994}   & 5794   \\
Stanford Cars            & 196                      & \multicolumn{1}{c|}{8144}   & 8041   \\
Stanford Online Products  & 22634                    & \multicolumn{1}{c|}{59551}  & 60502  \\
VehicleID (Full)         & 20207                    & \multicolumn{1}{c|}{113346} & 108221 \\
VehicleID (Small)         & 13603                    & \multicolumn{1}{c|}{113346} & 6493   \\
VehicleID (Medium)        & 14007                    & \multicolumn{1}{c|}{113346} & 13377  \\
VehicleID (Large)        & 14442                    & \multicolumn{1}{c|}{113346} & 19777  \\ \hline
\end{tabular}
}
\label{tab:stats-image-retrieval}
\end{table}


\subsection{Implementation Details}
\label{sec:implementation-details}
We follow the implementation described in \methodDEX{} for a fair comparison. Our base model is a ResNet-50. The explicit branch uses the cross-entropy (CE) identity classification loss. For the \methodDEX{} branch, we replace the CE loss with the \methodDEX{} loss as described in \autoref{eqn:dex}. For the reunification stage, we replace the metric triplet and center losses with our own contrastive-stream reunification versions as described in \autoref{sec:ablation-cross-stream-reunite-components}. We use a batch size of 32 which is duplicated for each of the two streams. We use a learning rate of $\eta = 1.75\text{e-}4$, starting at $0.01\eta$ and linearly warming up to $\eta$ in 10 epochs. We reduce the learning rate by a factor of 0.1 at epochs 30 and 55. We apply color jitter augmentation. For benchmarks with a large number of classes, such as the results presented in \autoref{tab:dg-reid-old}, we apply negative sampling of 2000 classes to fit the model into GPU memory~\citep{DEX}. 

For our semantic transformation policy, \PSTEAbrv{}, we sample a Gaussian model $\mathcal{N}$. We apply the expansion to all intermediate layers of the backbone: $L = [0,1,2,3,4]$. A minimum layer width of $\mu=3$ layers are considered for random selection at any time. Starting with $[0,1,2]$ in the first epoch, over $T=60$ epochs we progressively widen the set of candidate layers for selection. After $T=60$, we select layers from all the layers in $L$. 

For \DSDAbrv{}, we use equal weights $\beta_1=1, \: \beta_2=1$ between \methodDEX{} and CE losses. For \CSRAbrv{}, we apply relative weights of $\psi_1=2, \: \psi_2=5\text{e-}4, \: \psi_3=1$ for $L_{CSP}$, $L_{CSC}$ and $L_{CST}$ respectively. The model trains for a total of 500 epochs.

During inference, \methodDEX{} and \PSTEAbrv{} are disabled and the inputs feed forward through the backbone without semantic expansion.

\begin{table*}[ht]
\center
\caption{Results on the modern DG-ReID benchmarks. For the methods with the $\dag$, we evaluated the official open source implementation on this benchmark. \textbf{Bold} numbers are the best, while \underline{underlined} numbers are second.}
\resizebox{1.9\columnwidth}{!}{
\begin{tabular}{l|c|cc|cc|cc|cc|cc}
\hline
\multirow{2}{*}{Method} &
  \multirow{2}{*}{Training Data} &
  \multicolumn{2}{c|}{C+D+MS→M} &
  \multicolumn{2}{c|}{C+M+MS→D} &
  \multicolumn{2}{c|}{C+D+M→MS} &
  \multicolumn{2}{c|}{D+M+MS→C} &
  \multicolumn{2}{c}{\textbf{Average}}\\ \cline{3-12} 
          &            & Rank-1 & mAP  & Rank-1 & mAP  & Rank-1 & mAP  & Rank-1 & mAP & Rank-1 & mAP  \\ \hline
QAConv~\citep{QAConv}    & Train Only & 67.7   & 35.6 & 66.1   & 47.1 & 24.3   & 7.5  & 23.5   & 21.0 &  45.4  & 27.8 \\
META~\citep{META-DGReID} $\dag$  & Train Only & 66.7   & 44.6 & 61.8   & 42.7 & 32.4   & 13.1 & 21.3   & 21.6 &  45.6  & 30.5 \\
OSNet-IBN~\citep{OSNet} $\dag$ & Train Only & 73.4   & 45.1 & 61.5   & 42.3 & 35.7   & 13.7 & 20.9   & 20.9 &  47.9  & 30.5 \\
OSNet-AIN~\citep{OSNet} $\dag$ & Train Only & 74.2   & 47.4 & 62.7   & 44.5 & 37.9   & 14.8 & 22.4   & 22.4 &  49.3  & 32.3 \\
DualNorm~\citep{DualNorm} $\dag$  & Train Only & 78.9   & 52.3 & 68.5   & 51.7 & 37.9   & 15.4 & 28.0   & 27.6 &  53.3  & 36.8 \\
M$^3$L~\citep{M3L}    & Train Only & 75.9   & 50.2 & 69.2   & 51.1 & 36.9   & 14.7 & 33.1   & 32.1 &  53.8  & 37.0 \\
DEX~\citep{DEX}       & Train Only & 81.5   & 55.2 & \underline{73.7}   & 55.0 & 43.5   & 18.7 & 36.7   & 33.8 &  58.9  & 40.7 \\
ACL~\citep{ACL-DGReID} $\dag$      & Train Only & 82.8   & 58.4 & 71.7   & 53.4 & \underline{47.2}   & \underline{19.4} & 35.5   & 34.6 &  \underline{59.3}  & 41.5 \\ 
META~\citep{META-DGReID} $\dag$ & Train Only  & 66.7   & 44.6 & 61.8   & 42.7 & 32.4   & 13.1 & 21.3   & 21.6 & 45.6 & 30.5 \\
SIL~\citep{StyleInterleaved} $\dag$ & Train Only  & 79.2   & 52.8 & 68.3   & 47.0 &   36.9   & 13.8 & 29.5   & 28.9 & 53.5 & 35.6 \\
PAT~\citep{ni2023part} & Train Only   & 75.2   & 51.7 & 71.8   & 56.5 & 45.6   & \underline{21.6} & 31.1   & 31.5 & 55.9 & 40.3 \\
\hline \hline
CBN~\citep{CBN-ReID}       & Train+Test & 74.7   & 47.3 & 70.0   & 50.1 & 37.0   & 15.4 & 25.2   & 25.7 &  51.7  & 34.6 \\
SNR~\citep{SNR}       & Train+Test & 75.2   & 48.5 & 66.7   & 48.3 & 35.1   & 13.8 & 29.1   & 29.0 &  51.5  & 34.9 \\
MECL~\citep{MECL-ReID}      & Train+Test & 80.0   & 56.5 & 70.0   & 53.4 & 32.7   & 13.3 & 32.1   & 31.5 &  53.7  & 38.7 \\
RaMoE~\citep{RaMoE-Dai2021}     & Train+Test & 82.0   & 56.5 & 73.6   & \textbf{56.9} & 34.1   & 13.5 & 36.6   & 35.5 &  56.6  & 40.6 \\
MixNorm~\citep{MixNorm}   & Train+Test & 78.9   & 51.4 & 70.8   & 49.9 & \underline{47.2}   & \underline{19.4} & 29.6   & 29.0 &  56.6  & 37.4 \\
MetaBIN~\citep{choi2021metabin}   & Train+Test & \underline{83.2}   & \textbf{61.2} & 71.3   & 54.9 & 40.8   & 17.0 & \underline{38.1}   & \textbf{37.5} &  58.4  & \underline{42.7} \\ \hline
\textbf{\methodUDSX{}} (Ours) & Train Only & \textbf{85.7}   & \underline{60.4} & \textbf{74.7}   & \underline{55.8} & \textbf{47.6}   & \textbf{20.2} & \textbf{38.9}   & \underline{37.2} &  \textbf{61.7}  & \textbf{43.4} \\
\end{tabular}%
}
\label{tab:dg-reid-modern}
\end{table*}

\subsection{SOTA Methods Comparison}

\subsubsection{DG-ReID Task}
\label{sec:dg-reid-compare-results-trad}
\begin{table}[ht]
\center
\caption{Results on the traditional DG-ReID setting. To save space, methods are cited in~\ref{sec:dg-reid-compare-results-trad}}
\begin{adjustbox}{width=\linewidth}
\begin{tabular}{l|ccc|ccc}
\hline
\multirow{2}{*}{Method} & R-1 & R-5 & mAP & R-1 & R-5 & mAP \\ \cline{2-7} 
 & \multicolumn{3}{c|}{GRID} & \multicolumn{3}{c}{i-LIDS} \\ \hline
AugMining & 46.6 & 67.5 & - & 76.3 & 93.0 & - \\
DIR-ReID & 47.8 & 51.1 & 52.1 & 74.4 & 83.1 & 78.6 \\
RaMoE & 46.8 & - & 54.2 & 85.0 & - & 90.2 \\
SNR & 40.2 & - & 47.7 & 84.1 & - & 89.9 \\
DTIN-Net & 51.8 & - & 60.6 & 81.8 & - & 87.2 \\
MetaBIN & 48.4 & - & 57.9 & 81.3 & - & 87.0 \\
BCaR & 52.8 & - & - & 81.3 & - & - \\
\methodDEX{} & 53.3 & 69.4 & 61.7 & 86.3 & 95.2 & 90.7 \\
\textbf{\methodUDSX{}} & \textbf{56.9} & \textbf{72.7} & \textbf{63.0} & \textbf{86.7} & \textbf{98.0} & \textbf{91.4} \\ \hline
 & \multicolumn{3}{c|}{PRID} & \multicolumn{3}{c}{VIPeR} \\ \hline
AugMining & 34.3 & 56.2 & - & 49.8 & 70.8 & - \\
DIR-ReID & 71.1 & 82.4 & 75.6 & 58.3 & 66.9 & 62.9 \\
RaMoE & 57.7 & - & 67.3 & 56.6 & - & 64.6 \\
SNR & 52.1 & - & 66.5 & 52.9 & - & 61.3 \\
DTIN-Net & 71.0 & - & 79.7 & 62.9 & - & 70.7 \\
MetaBIN & 74.2 & - & 81.0 & 59.9 & - & 68.6 \\
BCaR & 70.2 & - & - & 65.8 & - & - \\
\methodDEX{} & 71.0 & 87.8 & 78.5 & 65.5 & 79.2 & 72.0 \\
\textbf{\methodUDSX{}} & \textbf{77.6} & \textbf{90.5} & \textbf{83.1} & \textbf{66.1} & \textbf{83.2} & \textbf{73.5} \\ \hline
\end{tabular}
\end{adjustbox}
\label{tab:dg-reid-old}
\end{table}

\methodUDSX{} outperforms or matches current SOTA methods in all standard evaluations for DG-ReID. \autoref{tab:dg-reid-modern} compares \methodUDSX{} with the recent strongest performing SOTA methods in the modern evaluation, which was established by M$^3$L~\citep{M3L}. As a test of \methodUDSX{}'s ability, we included results from other methods that tested on a closely-related benchmark where training and testing sets are merged together within the source domains to get more training data, while the target test domain remains the same. Even with the significant handicap of using less training data, our method is still able to surpass the performance of the methods using more data for a majority of the benchmarks, showing that \methodUDSX{} learns more efficiently from data than its competitors. If we restrict to only the same training-set-only benchmark, our method surpasses the SOTA in all evaluations by a significant margin. Furthermore, considering the average Rank-1 and mAP performance across all four target domains, our method still emerges on top regardless of amount of source training data, demonstrating the superior all-round performance of \methodUDSX{}.


\autoref{tab:dg-reid-old} presents \methodUDSX{} for the traditional evaluation methodology, again comparing with the strongest in the field such as AugMining~\citep{AugMining}, DIR-ReID~\citep{DIR-ReID}, RaMoE~\citep{RaMoE-Dai2021}, SNR~\citep{SNR}, DTIN-Net~\citep{DTIN-Net}, MetaBIN~\citep{choi2021metabin} and BCaR~\citep{BCaR-masato_bmvc2020}. In the modern evaluation, \methodUDSX{} surpasses the SOTA performance for all four benchmarks. Notably, \methodUDSX{}'s performance on C+D+MS $\rightarrow$ M pushed the SOTA by more than 2$\%$ in Rank-1 and mAP and for C+D+M $\rightarrow$ MS by more than 4$\%$ in Rank-1. In both modern and traditional DG-ReID benchmarks, \methodUDSX{} emerges in the first place and demonstrates the effectiveness of combining explicit and implicit semantic expansion methods in the task of DG-ReID.

\subsubsection{Image Retrieval Tasks}
\label{sec:image-retrieval-comparison}
\autoref{tab:object-retrieval-sota} compares \methodUDSX{} against other recent SOTA methods in image retrieval. For fairness, we compare methods that all use a ResNet-50 backbone, such as NormSoftMax~\citep{NormSoftMax}, MS512~\citep{MS512}, ROADMAP~\citep{ROADMAP}, ProxyNCA++~\citep{ProxyNCA++}, AVSL~\citep{AVSL}, Metrix~\citep{Metrix}, SCT~\citep{SCT}, Recall@k Surrogate Loss~\citep{RecallAtKSurrogate}, VehicleNet~\citep{VehicleNet}, Smooth-AP Loss~\citep{SmoothAPLoss} and RPTM~\citep{RPTM}. For completeness, we also compare with implicit semantic expansion techniques ISDA~\citep{ISDA-NeurIPS2019} and \methodDEX{}~\citep{DEX}.  In all of these benchmarks, \methodUDSX{} achieves the best Rank-1 score, demonstrating that our design improves implicit semantic expansion methods and allows for better performance not only in Person ReID but also in the image retrieval field at large.

\begin{table}[ht]
\center
\caption{Results on general image retrieval benchmarks. To save space, methods are cited in~\ref{sec:image-retrieval-comparison}}
\begin{adjustbox}{width=1\linewidth}
\begin{tabular}{lc|lccc} \hline
\multicolumn{1}{c|}{Method}          & Rank-1                & \multicolumn{1}{c|}{Method}          & \multicolumn{3}{c}{Rank-1}                                                                           \\ \hline
\multicolumn{2}{c|}{CUB-200-2011}                            & \multicolumn{4}{c}{Stanford Cars (Cars196)}                                                                                                 \\ \hline
\multicolumn{1}{l|}{NormSoftMax}     & 65.3                  & \multicolumn{1}{l|}{MS512}           & \multicolumn{3}{c}{84.1}                                                                             \\
\multicolumn{1}{l|}{MS512}           & 65.7                  & \multicolumn{1}{l|}{NormSoftmax}     & \multicolumn{3}{c}{89.3}                                                                             \\
\multicolumn{1}{l|}{ROADMAP}         & 68.5                  & \multicolumn{1}{l|}{ProxyNCA++}      & \multicolumn{3}{c}{90.1}                                                                             \\
\multicolumn{1}{l|}{ProxyNCA++}      & 72.2                  & \multicolumn{1}{l|}{AVSL}            & \multicolumn{3}{c}{91.5}                                                                             \\
\multicolumn{1}{l|}{ISDA}            & 73.4                  & \multicolumn{1}{l|}{ISDA}            & \multicolumn{3}{c}{90.9}                                                                             \\
\multicolumn{1}{l|}{\methodDEX{}}             & 75.4                  & \multicolumn{1}{l|}{\methodDEX{}}             & \multicolumn{3}{c}{91.2}                                                                             \\
\multicolumn{1}{l|}{\textbf{\methodUDSX{}}} & \textbf{77.8}         & \multicolumn{1}{l|}{\textbf{\methodUDSX{}}} & \multicolumn{3}{c}{\textbf{92.6}}                                                                    \\ \hline
\multicolumn{2}{c|}{Stanford Online Products}                & \multicolumn{4}{c}{VehicleID}                                                                                                               \\ \hline
\multicolumn{1}{l|}{}                & \multicolumn{1}{l|}{} & \multicolumn{1}{l|}{}                & \multicolumn{1}{l}{{\ul Small}} & \multicolumn{1}{l}{{\ul Medium}} & \multicolumn{1}{l}{{\ul Large}} \\
\multicolumn{1}{l|}{Metrix}          & 81.3                  & \multicolumn{1}{l|}{VehicleNet}      & 83.6                            & 81.4                             & 79.5                            \\
\multicolumn{1}{l|}{SCT}             & 81.6                  & \multicolumn{1}{l|}{Smooth-AP}       & 94.9                            & 93.3                             & 91.9                            \\
\multicolumn{1}{l|}{Recall@K}        & 82.7                  & \multicolumn{1}{l|}{RPTM}            & 95.1                            & 93.3                             & 92.7                            \\
\multicolumn{1}{l|}{ROADMAP}         & 83.1                  & \multicolumn{1}{l|}{Recall@K}        & 95.7                            & 94.6                             & 93.8                            \\
\multicolumn{1}{l|}{ISDA}            & 83.4                  & \multicolumn{1}{l|}{ISDA}            & 94.7                            & 93.2                             & 92.8                            \\
\multicolumn{1}{l|}{\methodDEX{}}             & 83.5                  & \multicolumn{1}{l|}{\methodDEX{}}             & 94.2                            & 94.2                             & 92.2                            \\
\multicolumn{1}{l|}{\textbf{\methodUDSX{}}} & \textbf{85.0}         & \multicolumn{1}{l|}{\textbf{\methodUDSX{}}} & \textbf{96.8}                   & \textbf{95.0}                    & \textbf{94.7}                   \\ \hline
\end{tabular}
\end{adjustbox}
\label{tab:object-retrieval-sota}
\end{table}

\subsection{Ablation Studies}
\subsubsection{Components of \methodUDSX{}}
\label{sec:ablation-components}
\autoref{tab:ablation-components} presents an ablation study of the influence due to the components of \methodUDSX{}. The study is performed on two DG-ReID benchmarks (C+D+MS $\rightarrow$ M and C+D+M $\rightarrow$ MS) and one general image retrieval benchmark (CUB-200-2011). The first two configurations, \PSTEAbrv{}/\methodDEX{} Only, do not require duplicate data streams, meaning that \DSDAbrv{} is switched off. In \PSTEAbrv{} Only, we perform explicit semantic expansion following the scheme described in \autoref{subsubsec:progressive-stochastic-semantic-expansion} and use the standard cross-entropy, center and triplet losses while leaving out \methodDEX{} and \CSRAbrv{}. Similarly, in \methodDEX{} Only we train only with the \methodDEX{} loss and omit everything else. \methodDEX{}+Naive performs explicit semantic expansion on a single stream of features while also applying the \methodDEX{} loss. It demonstrates that naively combining them leads to poor results. For the rest of the configurations, we cumulatively add components of \methodUDSX{} to \methodDEX{}, starting with \DSDFull{} (\methodDEX{} + \DSDAbrv{}), followed by \PSTEFull{} (\methodDEX{} + \DSDAbrv{} + \PSTEAbrv{}) and finally add \CSRFull{} to get our full method, \methodUDSX{}. Our experiments demonstrate that each component improves generalization scores in DG-ReID and also in image retrieval. Furthermore, we observe that using PSTE Only surpasses DEX Only in some benchmarks like C+D+MS $\rightarrow$ M and CUB-200-2011, but their combined strengths yield the best results.
\begin{table}[ht]
\caption{Ablation over configurations of \methodUDSX{}. For \PSTEAbrv{}/\methodDEX{} Only, we only use a single data stream and train with the respective loss. Only configurations with \DSDAbrv{} require two data streams.}
\begin{adjustbox}{width=1\linewidth}
\begin{tabular}{c|l|cc} \hline
Benchmark                              & Configuration                           & Rank-1 & mAP  \\ \hline
\multirow{5}{*}{C+D+MS $\rightarrow$ M}  & \PSTEAbrv{} Only                         & 81.8   & 55.5 \\
                                        & \methodDEX{} Only                           & 80.0   & 55.1 \\
                                        & \methodDEX{} + Naive                         & 79.8   & 53.6 \\
                                        & \methodDEX{} + \DSDAbrv{}                       & 84.0   & 58.2 \\
                                        & \methodDEX{} + \DSDAbrv{} + \PSTEAbrv{}     & 84.8   & 58.4 \\
                                        & \textbf{\methodUDSX{} (Ours)} & \textbf{85.7}   & \textbf{60.4} \\ \hline
\multirow{5}{*}{C+D+M $\rightarrow$ MS} & \PSTEAbrv{} Only                            & 40.9   & 16.5 \\
                                        & \methodDEX{} Only                            & 43.0   & 18.3 \\
                                        & \methodDEX{} + Naive                         & 43.2   & 18.2 \\
                                        & \methodDEX{} + \DSDAbrv{}                       & 44.4   & 19.1 \\
                                        & \methodDEX{} + \DSDAbrv{} + \PSTEAbrv{}     & 45.7   & 19.4 \\
                                        & \textbf{\methodUDSX{} (Ours)} & \textbf{47.6}   & \textbf{20.2} \\ \hline
\multirow{5}{*}{CUB-200-2011}                    & \PSTEAbrv{} Only                            & 74.2   & 61.9 \\
                                        & \methodDEX{} Only                            & 75.8   & 60.9 \\
                                        & \methodDEX{} + Naive                         & 75.4   & 54.7 \\
                                        & \methodDEX{} + \DSDAbrv{}                       & 76.5   & 62.5 \\
                                        & \methodDEX{} + \DSDAbrv{} + \PSTEAbrv{}     & 77.1   & 63.1 \\
                                        & \textbf{\methodUDSX{} (Ours)} & \textbf{77.6}   & \textbf{63.6} \\ \hline
\end{tabular}
\end{adjustbox}
\label{tab:ablation-components}
\end{table}

\subsubsection{Hyperparameter Study}
\label{sec:hyperparameter-study}
Our method \methodUDSX{} employs five important parameters that belong in two groups. The first group, $\beta_1$ and $\beta_2$, control the strengths of explicit and implicit semantic expansion, while the second group, $\psi_1$, $\psi_2$ and $\psi_3$, balance the components in the reunification process of \CSRFull{}. We investigate the effects of these two groups of hyperparameters in each of the following sub-sections, using the same two multi-source DG-ReID benchmarks, C+D+MS $\rightarrow$ M and C+D+M $\rightarrow$ MS.

\subsubsection{Explicit/Implicit Trade-off}
Table~\ref{tab:explicit-implicit-tradeoff} compares the effects of varying the relative strengths between explicit ($\beta_1$) and implicit ($\beta_2$) semantic expansion. Starting with $\beta_1=1, \beta_2=0$, we increase the strength of \methodDEX{} via $\beta_2$ until they are evenly matched at $\beta_1=1, \beta_2=1$. We do this in the opposite direction as well, starting from $\beta_1=0, \beta_2=1$ and increasing $\beta_1$ until we reach $\beta_1=1, \beta_2=1$. First, we can observe that when operating individually without the other, implicit semantic expansion is stronger than the explicit by a few percentage points in Rank-1 and mAP for both benchmarks. Furthermore, even with a relatively low implicit semantic expansion weight ($\beta_2$), the performance of our \methodUDSX{} already increases significantly. However, we also observe that further increasing $\beta_2$ beyond this low weight does not improve performance much. Instead, while explicit semantic expansion alone is comparatively less effective, it plays a strong supporting role by boosting the performance beyond the limitations imposed by the implicit semantic expansion, as discussed in~\autoref{sec:analysis}. This comparison experimentally justifies our intention to supplement \methodDEX{} with explicit semantic expansion to help overcome its limits.

\subsubsection{CSR Components Analysis}
\label{sec:ablation-cross-stream-reunite-components}
We study the effects of the three components of CSR, as shown in~\autoref{tab:ablation-csr}. For the \CSPAbrv{} loss study, we compare a baseline without \CSPAbrv{} Loss against three other models trained on \CSCAbrv{} loss with weights $\psi_1 \in \{1,2,5\}$. \CSCAbrv{} and \CSTAbrv{} are compared against the standard Center~\cite{Wen2016ARecognition} and Triplet~\cite{TriNet} losses respectively with the same weight. For \CSCAbrv{}/Center loss, the standard weight is $\psi_2=5\text{e-}4$ and for \CSTAbrv{}/Triplet loss it is $\psi_3=1$. Our experiment shows that for \CSPAbrv{} loss, a moderate weight of $\psi_1=1$ improves results. Also, we found that replacing the standard Center and Triplet losses with our \CSCAbrv{} and \CSTAbrv{} losses improves results significantly. These experiments show that the \CSPFull{} is an effective stream unifier and the other components of \CSRFull{} are more effective over the standard Center and Triplet losses.

\begin{table}[ht]
\caption{Varying the strengths between explicit (\PSTEAbrv{}) and implicit (\methodDEX{}) semantic expansion, both of which are controlled by hyperparameters $\beta_1$ and $\beta_2$ respectively (See Eq.~\ref{eq:nd-loss}).}
\begin{adjustbox}{width=1\linewidth}
\begin{tabular}{c|cc|cc}
\hline
Benchmark                                 & \multicolumn{1}{c}{$\beta_1$} & \multicolumn{1}{c|}{$\beta_2$} & \multicolumn{1}{c}{Rank-1} & \multicolumn{1}{c}{mAP} \\ \hline
\multirow{7}{*}{C+D+MS $\rightarrow$ M}
 & 1.0 & 0.0    & 81.7 & 54.1  \\
 & 1.0 & 0.4    & 84.9 & 59.9 \\
 & 1.0 & 0.8    & 85.1 & 59.9 \\
 & 1.0 & 1.0    & \textbf{85.7} & \textbf{60.4}  \\
 & 0.8 & 1.0    & 85.6 & 60.1 \\
 & 0.4 & 1.0    & 85.0 & 59.6 \\
 & 0.0   & 1.0  & 83.0 & 56.5 \\ \hline
\multirow{7}{*}{C+D+M $\rightarrow$ MS}
 & 1.0 & 0.0   & 41.1 & 16.7 \\
 & 1.0 & 0.4 & 44.8 & 18.9 \\
 & 1.0 & 0.8 & 45.6 & 19.2 \\
 & 1.0 & 1.0 & \textbf{47.6} & \textbf{20.2} \\
 & 0.8 & 1.0 & 45.9 & 19.6 \\
 & 0.4 & 1.0 & 45.0 & 19.1 \\
 & 0.0   & 1.0 & 45.0 & 19.0 \\ \hline
\end{tabular}
\end{adjustbox}
\label{tab:explicit-implicit-tradeoff}
\end{table}

\begin{table}[ht]
\caption{Ablation study over components of the contrastive stream losses in \CSRAbrv{} (See Eq.~\ref{eq:unification-loss}).}
\begin{adjustbox}{width=1\linewidth}
\begin{tabular}{c|l|cc}
\hline
Benchmark & \multicolumn{1}{c|}{Config} & Rank-1 & mAP \\ 
\hline
\multirow{8}{*}{C+D+MS → M} 
& Baseline & 81.7 & 54.7 \\
& \CSPAbrv{} ($\psi_1=1$) & \textbf{85.4} & \textbf{59.0} \\
& \CSPAbrv{} ($\psi_1=2$) & 84.9 & 59.0 \\
& \CSPAbrv{} ($\psi_1=5$) & 85.1 & 58.9 \\ \cline{2-4} 
& Center ($\psi_2=5\text{e-}4$) & 80.8 & 54.0 \\
& \CSCAbrv{} ($\psi_2=5\text{e-}4$) & \textbf{84.3} & \textbf{57.8} \\ \cline{2-4} 
& Triplet ($\psi_3=1$) & 81.9 & 55.8 \\
& \CSTAbrv{} ($\psi_3=1$) & \textbf{85.1} & \textbf{57.9} \\ \hline
\multirow{8}{*}{C+D+M → MS} 
& Baseline & 43.3 & 17.9 \\
& \CSPAbrv{} ($\psi_1=1$) & \textbf{43.7} & \textbf{18.4} \\
& \CSPAbrv{} ($\psi_1=2$) & 42.9 & 17.7 \\
& \CSPAbrv{} ($\psi_1=5$) & 42.3 & 17.7 \\ \cline{2-4} 
& Center ($\psi_2=5\text{e-}4$) & 43.4 & 18.7 \\
& \CSCAbrv{} ($\psi_2=5\text{e-}4$) & \textbf{45.9} & \textbf{19.0} \\ \cline{2-4} 
& Triplet ($\psi_3=1$) & 43.0 & 18.0 \\
& \CSTAbrv{} ($\psi_3=1$) & \textbf{43.6} & \textbf{18.5} \\ \hline
\end{tabular}
\end{adjustbox}
\label{tab:ablation-csr}
\end{table}

\subsubsection{Robustness to Over-Fitting}
\label{sec:ablation-robustness-lambda}
Table~\ref{tab:lambda-comparison-after} demonstrates that \methodUDSX{} is able to achieve good results under stronger application of \methodDEX{}, which is controlled by $\lambda$. We compared \methodUDSX{} against \methodDEX{} across the four major DG-ReID benchmarks and across values of $\lambda \in \{5,15,25,50\}$. Results for \methodUDSX{} are presented first while results for \methodDEX{} are in parentheses. Across the board, the addition of explicit semantic expansion is shown to yield a net positive benefit. Compared with \methodDEX{} alone, we can see that the epoch at which model performance peaks (Best Epoch) is significantly higher for \methodUDSX{}. In C+D+MS $\rightarrow$ M and D+M+MS $\rightarrow$ C, the application of explicit semantic expansion shows a clear and significant performance improvement, while in C+D+M $\rightarrow$ MS explicit expansion improves Rank-1 by a wide margin. In these benchmarks, we demonstrate that \methodUDSX{} mitigates the degradation of performance caused by over-fitting, allowing the model to train effectively for more epochs and reach better evaluation scores than before.
\begin{table}[ht]
    \centering
    \caption{Varying strength of DEX via $\lambda$ on all four major DG-ReID benchmarks. Values in parentheses are for \methodDEX{} only with no explicit expansion. Through this study, we demonstrate that our design choices in \methodUDSX{} mitigate early over-fitting and can accommodate stronger \methodDEX{} weights.}
    \begin{adjustbox}{width=1\linewidth}
\begin{tabular}{c|c|c|cc}
\hline
Benchmark                               & $\lambda$          & Best Epoch & mAP           & Rank-1        \\ \hline
\multirow{5}{*}{C+D+MS $\rightarrow$ M} & No DEX ($\lambda=0$) & 307 (57)        & 55.1 (54.0)          & 82.1 (79.8)          \\
                                        & 5               & 310 (39)        & 59.2 (54.8)         & 85.1 (81.1)         \\
                                        & 15     & 322 (39)        & \textbf{60.4} (54.2) & \textbf{85.7} (80.2) \\
                                        & 25              & 322 (35)        & 59.7 (54.1)          & 85.0 (79.7)         \\
                                        & 50              & 112 (34)       & 59.3 (52.7)          & 84.7 (78.1)         \\ \hline
\multirow{5}{*}{C+D+M $\rightarrow$ MS} & No DEX ($\lambda=0$) & 68 (201)         & 16.5 (18.3)          & 40.9 (43.4)         \\
                                        & 5               & 156 (99)        & 19.3 (18.6)          & 45.4 (43.9)          \\
                                        & 15     & 198 (99)        & \textbf{20.2} (19.1) & \textbf{47.6} (44.8) \\
                                        & 25              & 198 (114)       & 19.3 (18.5)          & 45.8 (43.0)         \\
                                        & 50              & 60 (45)         & 19.0 (18.5)          & 44.6 (43.3)          \\ \hline
\multirow{5}{*}{C+M+MS $\rightarrow$ D} & No DEX ($\lambda=0$) & 391 (252)         & 53.8 (54.1)          & 72.3 (72.6)          \\
                                        & 5               & 382 (105)        & 54.0 (54.6)         &   72.1 (72.3)        \\
                                        & 15     & 287 (78)        & \textbf{55.8} (54.9) & \textbf{74.7} (73.5) \\
                                        & 25              & 238 (72)        & 55.0 (54.0)          & 73.8 (73.0)          \\
                                        & 50              & 307 (42)         & 54.9 (53.7)         & 73.1 (72.5)          \\ \hline
\multirow{5}{*}{D+M+MS $\rightarrow$ C} & No DEX ($\lambda=0$) & 492 (111)         & 34.2 (29.3)          & 35.2 (29.5)         \\
                                        & 5               & 357 (120)        & 34.7 (30.4)         & 36.2 (31.6)         \\
                                        & 15     & 207 (108)        & \textbf{37.2} (33.2) & \textbf{38.9} (34.1) \\
                                        & 25              & 265 (114)        & 36.5 (33.8)          & 37.1 (35.4)          \\
                                        & 50              & 137 (78)         & 35.2 (33.5)          & 35.4 (35.1)         \\ \hline
\end{tabular}    \end{adjustbox}
    \label{tab:lambda-comparison-after}
\end{table}
\section{Conclusion}
Implicit semantic expansion methods, while powerful, have inherent drawbacks that limit its generalization capacity when learning from datasets with a large number of classes. Naively combining implicit and explicit semantic expansion is generally detrimental and requires finesse in order to get them to work together. In this work, we unified both of them in a single framework, \methodUDSX{}, such that the strengths of both types of semantic expansion are synergized, and achieve new state-of-the-art results in DG-ReID and general image retrieval benchmarks.

\section*{Acknowledgement}
This work was supported by the Defence Science and Technology Agency (DSTA) Postgraduate Scholarship, of which Eugene P.W. Ang is a recipient. It was carried out at the Rapid-Rich Object Search (ROSE) Lab at the Nanyang Technological University, Singapore.
\printcredits

\section*{Declaration of competing interest}
The authors declare that they have no known competing financial interests or personal relationships that could have appeared to influence the work reported in this paper.

\section*{Data availability}
Data will be made available on request.

\bibliographystyle{model1-num-names}

\pagebreak
\bio{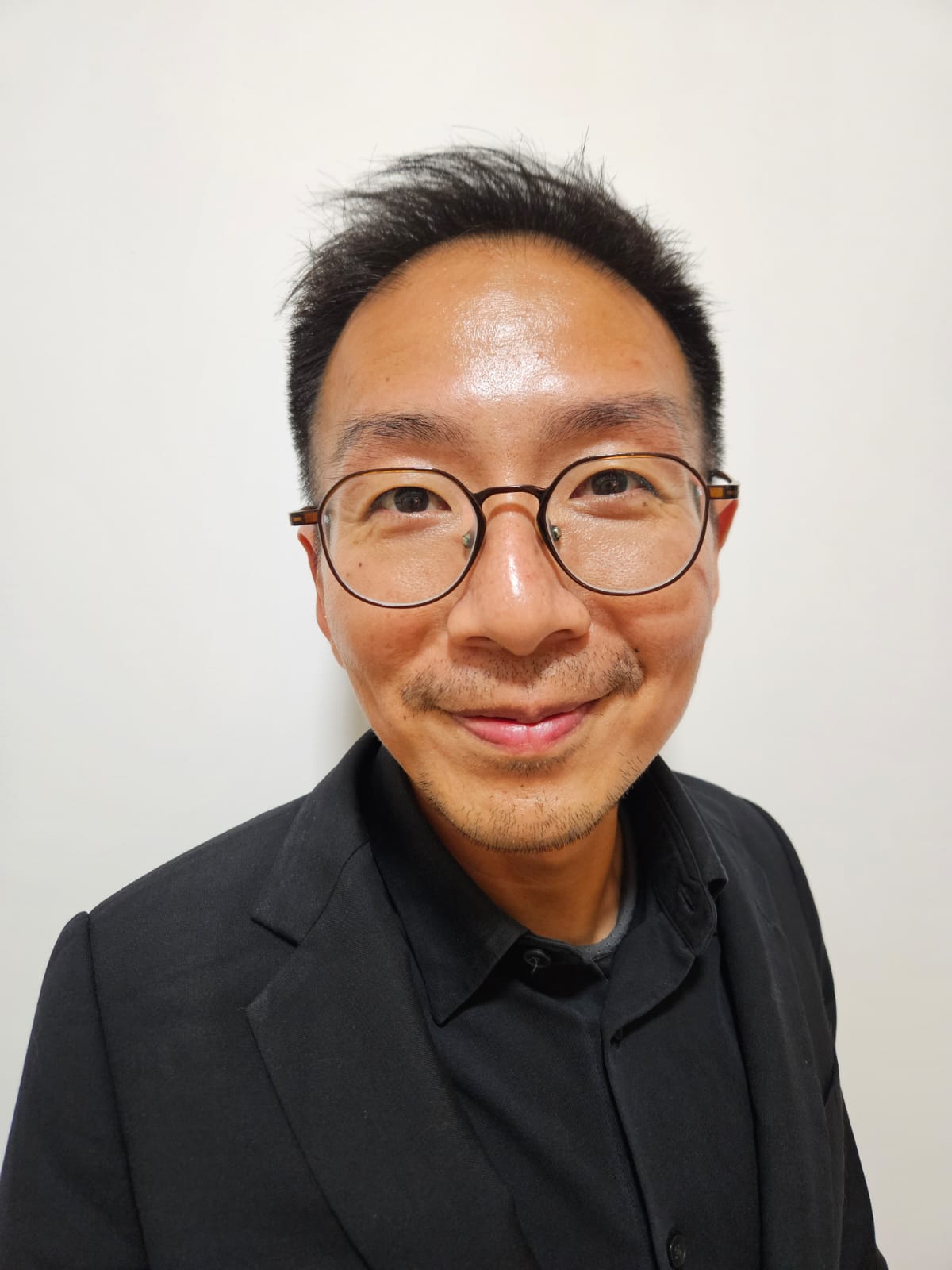}
Eugene P.W. Ang received both the B.Sc. in Computer Science in 2007 and M.Sc. in Information Systems Management in 2008 from Carnegie Mellon University, USA. He also received the M.Sc. in Computer Science (Machine Learning Track) from Columbia University, USA, in 2017. Under a post-graduate scholarship from the Defence Science and Technology Agency he is currently a PhD candidate in ROSE Lab, Nanyang Technological University, Singapore, where he pursues research on the practical applications of deep learning and computer vision such as domain generalized person/object re-identification and has published several papers in these areas.
\endbio

\bio{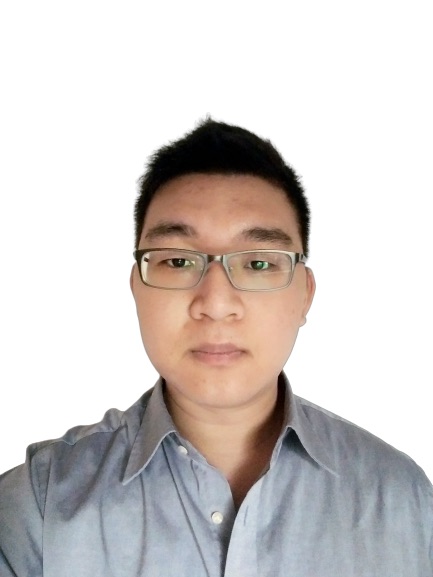}
Shan Lin received the B.Sc. degree and Ph.D. degree from the University of Warwick, U.K, in 2015 and 2020. He is currently a research fellow in ROSE Lab, Nanyang Technological University, Singapore. His current research interests are in the area of person re-identification, computer vision, and deep learning. His studies are funded by the European Union EU H2020 project IDENTITY and National Research Foundation, Singapore under AI Singapore program. He has published several technical papers in these areas.
\endbio

\bio{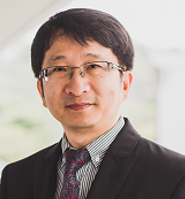}
Prof. Alex C. Kot has been with the Nanyang Technological University (NTU), Singapore since 1991. He headed the Division of Information Engineering at the School of Electrical and Electronic Engineering (EEE) for eight years. He was the Vice Dean Research and Associate Chair (Research) for the School of EEE for three years, overseeing the research activities for the School with over 200 faculty members. He was the Associate Dean (Graduate Studies) for the College of Engineering (COE) for eight years. He is currently the Director of ROSE Lab [Rapid(Rich) Object SEearch Lab) and the Director of NTU-PKU Joint Research Institute. He has published extensively with over 300 technical papers in the areas of signal processing for communication, biometrics recognition, authentication, image forensics, machine learning and AI. Prof. Kot served as Associate Editor for a number of IEEE transactions, including IEEE TSP, IMM, TCSVT, TCAS-I, TCAS-II, TIP, SPM, SPL, JSTSP, JASP, TIFS, etc. He was a TC member for several IEEE Technical Committee in SPS and CASS. He has served the IEEE in various capacities such as the General Co-Chair for the 2004 IEEE International Conference on Image Processing (ICIP) and area/track chairs for several IEEE flagship conferences. He also served as the IEEE Signal Processing Society Distinguished Lecturer Program Coordinator and the Chapters Chair for IEEE Signal Processing Chapters worldwide. He received the Best Teacher of The Year Award at NTU, the Microsoft MSRA Award and as a co-author for several award papers. He was elected as the IEEE CAS Distinguished Lecturer in 2005. He was a Vice President in the Signal Processing Society and IEEE Signal Processing Society Distinguished Lecturer. He is now a Fellow of the Academy of Engineering, Singapore, a Life Fellow of IEEE and a Fellow of IES.
\endbio

\end{sloppypar}
\end{document}